\documentclass[10pt,twocolumn,letterpaper]{article}
\usepackage{wacv}          
\usepackage{xcolor}

\definecolor{myblue}{HTML}{0070C0}
\newcommand{\bblue}[1]{\textcolor{myblue}{#1}}
\definecolor{oorange}{HTML}{DD8047}
\newcommand{\oorange}[1]{\textcolor{oorange}{#1}}
\definecolor{ggreen}{HTML}{27ae60}
\newcommand{\ggreen}[1]{\textcolor{ggreen}{#1}}
\definecolor{slotblue}{RGB}{40,105,180}

\usepackage{tabularx}
\usepackage{booktabs}
\usepackage{multirow}
\usepackage{array}
\usepackage{ragged2e}

\usepackage{listings}

\lstdefinestyle{jsonprompt}{
    basicstyle=\footnotesize\ttfamily,
    columns=fullflexible,
    keepspaces=true,
    breaklines=true,
    breakatwhitespace=false,
    showstringspaces=false,
    escapeinside={(*@}{@*)},
    aboveskip=2pt,
    belowskip=2pt
}

\usepackage[table]{xcolor}

\usepackage{graphicx}
\usepackage{amssymb}

\definecolor{ours}{RGB}{235,245,235}

\usepackage[normalem]{ulem} 

\definecolor{bestblue}{RGB}{0,114,178}
\definecolor{secondorange}{RGB}{213,94,0}

\newcommand{\best}[1]{{\textcolor{bestblue}{#1}}}
\newcommand{\second}[1]{{\textcolor{secondorange}{#1}}}

\definecolor{wacvblue}{rgb}{0.21,0.49,0.74}
\usepackage[pagebackref,breaklinks,colorlinks,allcolors=wacvblue]{hyperref}

\title{Open-Attribute Person Retrieval:\\Finding People Through Distinctive and Novel Attributes}

\author{Minjeong Park$^{1*}$ \qquad
Hongbeen Park$^{1*}$ \qquad
Sangwon Lee$^2$ \qquad
Jinkyu Kim$^{3,4}$
\\
$^{1}$The Pennsylvania State University \quad
$^{2}$Korea Telecom Research \quad
$^{3}$Korea University \quad
$^{4}$Kakao Mobility \\
{\tt\small \{minjeongpark,hongbeenpark\}@psu.edu}\\
{\tt\small lee.sangwon@kt.com \quad jinkyukim@korea.ac.kr}
}

\begin{document}
\maketitle
\begingroup
\renewcommand\thefootnote{}
\footnotetext{* This work was done while the authors were at Korea University.}
\endgroup

\begin{abstract}
Person retrieval in surveillance videos often depends on attributes described by witnesses or operators. However, the most useful cues in practice are not always common appearance descriptions (e.g., gender, clothing color), but rare and distinctive attributes that can sharply reduce the search space (e.g., holding a weapon, lying on the ground). Existing text-based person retrieval benchmarks and methods largely focus on identity-centric retrieval with common pedestrian descriptions, leaving such retrieval-critical attributes underexplored. In this paper, we introduce Open-Attribute Person Retrieval (OAPR), a practical retrieval setting that aims to retrieve all pedestrian instances matching a given attribute query, including rare or previously unseen visual concepts, regardless of identity. To support this task, we construct EPAD, an Expanded Pedestrian Attribute Dataset with 267,885 pedestrian images and a unified vocabulary of 65 attributes, including safety-critical actions, assistive devices, and object interactions that are rarely covered in prior benchmarks. We further propose GAP-CLIP, a lightweight CLIP-based framework that learns gated attribute-aware body-part representations for OAPR. Extensive experiments on EPAD demonstrate that GAP-CLIP achieves the strongest top-K retrieval performance on the full attribute space and on out-of-distribution attributes. The code and dataset are available at \url{https://github.com/mlnjeongpark/Open-Attribute-Person-Retrieval}.
\vspace{-1.5em}
\end{abstract}
    
\section{Introduction}
\label{sec:intro}
\begin{figure}[t]
    \centering
    \includegraphics[width=\linewidth]{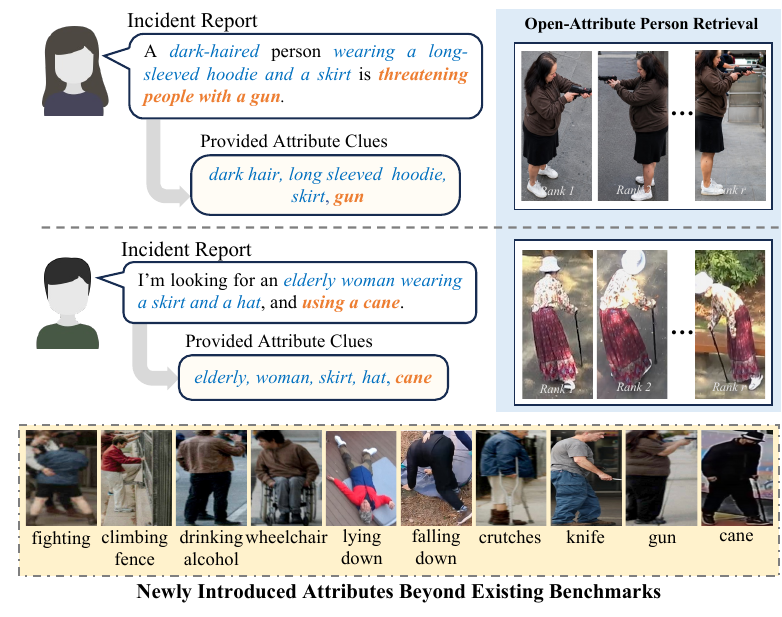}
    \caption{
    \textbf{(Top)}
Open-Attribute Person Retrieval moves beyond identifying people based on \bblue{common visual descriptions shared across many pedestrians} and instead focuses on retrieving individuals using \oorange{distinctive features that facilitate person differentiation but are not present in the benchmarks}. \textbf{(Bottom)} We further introduce EPAD, a new benchmark with additional attribute annotations absent from existing pedestrian attribute datasets.
    }
    \label{fig:teaser}
\vspace{-1.5em}
\end{figure}

Searching for a person-of-interest in surveillance videos often requires reasoning over the visual cues described by a witness or an operator. When a surveillance operator receives a report of a person-of-interest, the most effective search cue is often not a common appearance description shared by many pedestrians, but a distinctive attribute that sharply narrows the candidate set. For example, searching for a ``dark-haired person wearing a long-sleeved hoodie and a skirt'' may return many candidates in a crowded scene. However, if the report also mentions that the person is holding a gun, this rare and highly discriminative cue can substantially reduce the search space. This suggests that practical person search should not only rely on common demographic or clothing descriptions, but also recognize distinctive attributes that are critical for identifying the target.

Despite their practical importance, such retrieval-critical attributes are largely underexplored in existing text-based person retrieval benchmarks. Most existing descriptions~\cite{PA100k,PETA,RAP2,jin2024pedestrian,li2017person,ding2021semantically,zhu2021dssl} focus on common demographic, clothing, and accessory cues, while assistive devices (e.g., cane, crutches) and threat-related object interactions (e.g., firearm, knife) are rarely annotated. Moreover, conventional text-based person retrieval is typically formulated as an identity-centric task, where the goal is to retrieve images of the same person described by a sentence. However, in surveillance scenarios, an operator may need to retrieve any pedestrian satisfying a given attribute query, regardless of identity. For example, pedestrians who are holding a weapon or lying on the street may need to be retrieved immediately regardless of their identity.

To address this challenge, we introduce \textbf{Open-Attribute Person Retrieval}, a task that aims to retrieve pedestrians based on attribute queries that may include rare or previously unseen visual concepts. Unlike conventional identity-centric retrieval, the goal is to retrieve all pedestrian instances that match the given attribute description, regardless of their identity. This setting requires models to recognize not only common attributes observed during training, but also open-ended attributes that may be absent or underrepresented in existing pedestrian datasets. 

To support this task, we construct \textbf{EPAD}, an \textbf{E}xpanded \textbf{P}edestrian \textbf{A}ttribute \textbf{D}ataset, which unifies existing pedestrian attribute datasets and extends their annotation space with retrieval-critical attributes. EPAD defines a unified vocabulary of 65 pedestrian attributes, including attributes that are rarely covered in prior benchmarks, such as \emph{fighting}, \emph{climbing fence}, \emph{wheelchair}, \emph{drinking alcohol}, \emph{gun}, \emph{knife}, \emph{cane}, \emph{lying down}, \emph{crutches}, and \emph{falling down}. EPAD contains 267,885 pedestrian images collected from existing pedestrian attribute datasets, surveillance anomaly data, and attribute-targeted synthetic samples. During dataset construction, we manually annotate missing or newly introduced attributes, resulting in approximately 4M binary attribute annotations.

Building on EPAD, we propose \textbf{GAP-CLIP}, a CLIP-based framework that learns \textbf{G}ated \textbf{A}ttribute-aware body-\textbf{P}art representations for Open-Attribute Person Retrieval. While vision-language models provide useful open-vocabulary knowledge, directly applying them to surveillance pedestrian images remains challenging due to the domain gap, low image resolution, and the fine-grained nature of pedestrian attributes. For example, some attributes can be inferred from global body appearance (e.g., age, gender), whereas others require attention to small local regions (e.g., accessories, assistive devices).
To address these challenges, GAP-CLIP leverages the open-vocabulary knowledge of a lightweight vision-language model while enhancing its ability to capture local pedestrian cues through body-part-aware representations. Specifically, it generates pseudo body features, learns body-part prompts, and selects attribute-relevant body representations through attribute-guided attention. 
Furthermore, a gated residual connection preserves the original CLIP class-token representation as a zero-shot anchor, preventing base-attribute training from overly distorting the pre-trained vision-language embedding space.

Our contributions are summarized as follows:
\begin{itemize}
    \item We introduce Open-Attribute Person Retrieval, a practical retrieval setting that requires retrieving all pedestrians matching an attribute query, including rare and previously unseen attributes. 
    
    \item We construct EPAD, a large-scale OAPR benchmark with 267,885 pedestrian images and a unified vocabulary of 65 attributes, including retrieval-critical attributes that are underrepresented in prior pedestrian benchmarks. 

    \item We propose GAP-CLIP, a lightweight vision-language framework for OAPR. GAP-CLIP learns body-part-aware pedestrian representations, selects attribute-relevant local cues through attribute-guided attention, and preserves CLIP's zero-shot knowledge via a gated residual connection.

    \item Comprehensive experiments on EPAD demonstrate that GAP-CLIP achieves superior performance on both base and novel attributes, showing its effectiveness for open-attribute person retrieval in practical surveillance scenarios. More results can be found in the supplementary material.
\end{itemize}





\begin{figure*}[t]
    \centering
    \includegraphics[width=\textwidth]{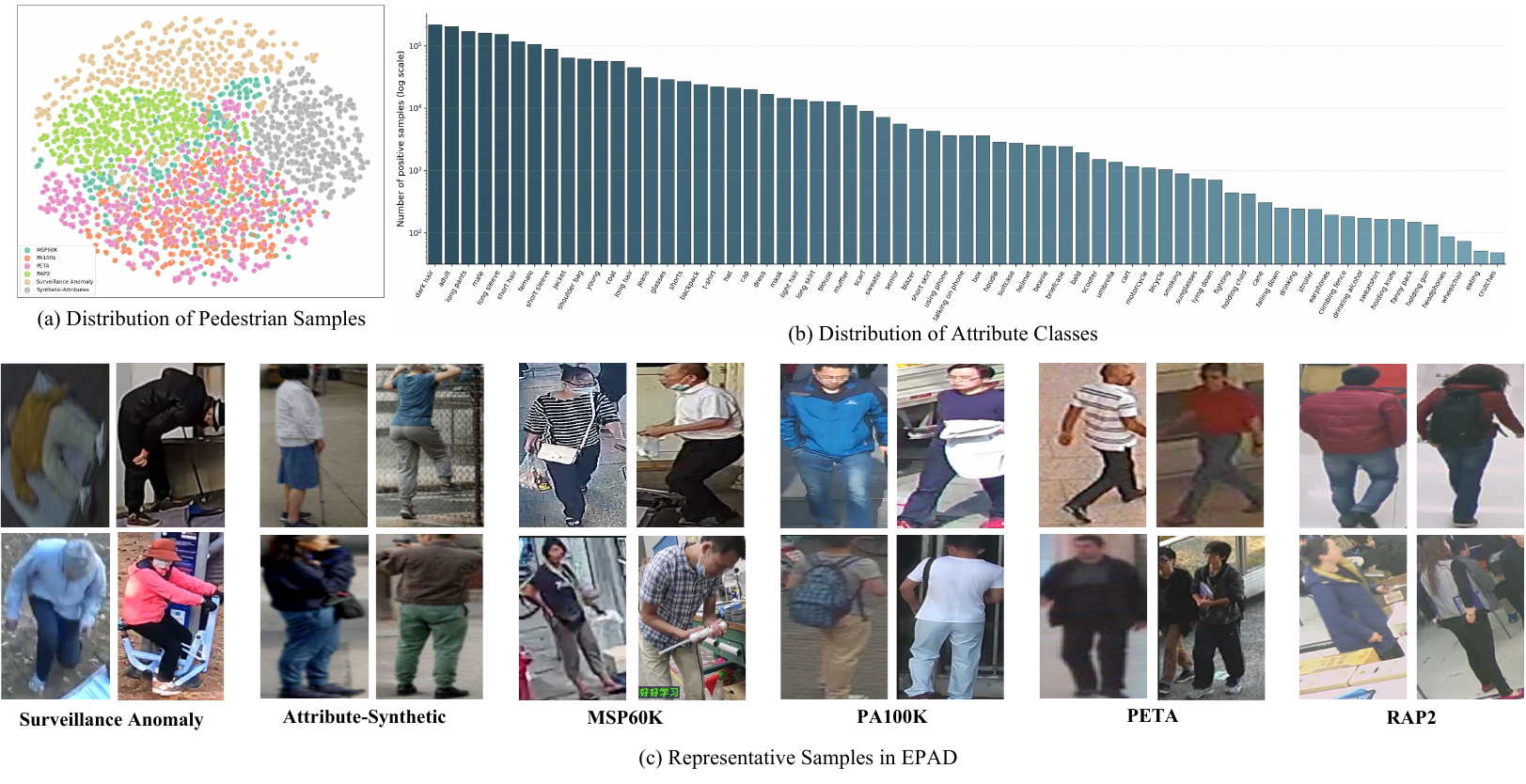}
\caption{
Overview of EPAD statistics. EPAD combines diverse pedestrian domains and expands the attribute space with rare but retrieval-critical cues such as abnormal actions, assistive devices, and object interactions.
}
\vspace{-1em}
    \label{fig:dataset}
\end{figure*}

\section{Expanded Pedestrian Attribute Dataset (EPAD)}
In this section, we present the Expanded Pedestrian Attribute Dataset (EPAD), a new benchmark for attribute-based person retrieval that retrieves pedestrians from textual attribute descriptions. Unlike existing pedestrian benchmarks that primarily cover common appearance attributes, EPAD substantially expands the attribute space with retrieval-critical cues. 
The detailed dataset collection and annotation procedures are described in the following sections.

\subsection{Pedestrian Data Collection}
While existing pedestrian attribute datasets provide diverse annotations for common visual cues, the images mostly capture ordinary walking or standing scenarios, with limited visual instances of rare but safety-critical cases (e.g., lying down, climbing a fence, or holding a weapon). To address this scarcity, we generate attribute-targeted synthetic pedestrian images using an image generation model and construct Attribute-Synthetic as a supplementary dataset for EPAD. Furthermore, we incorporate surveillance anomaly datasets~\cite{AIHub-2,AIHub-multi,AIHub-senior} to obtain more diverse visual samples of rare safety-critical scenarios that are underrepresented in conventional pedestrian datasets.
 Our pedestrian data comes from three sources:

\noindent\textbf{(1) Attribute-Synthetic:} 
To construct Attribute-Synthetic, we use a structured JSON-based prompt generation pipeline for Ideogram 4.0~\cite{ideogram-4-2026}. 
We generate 200 samples for each behavior category covering \textit{climbing a fence, drinking alcohol, fighting, holding a cane, holding a child, holding a gun, holding a knife, lying on the ground, riding a bicycle, smoking, using crutches,} and \textit{using a wheelchair}. To better approximate real-world surveillance imagery, we further apply post-processing operations, including Gaussian noise, motion blur, JPEG compression, brightness and contrast adjustment, gamma correction, and desaturation. Finally, we perform manual quality control to remove generation failures, such as images with severe artifacts, incorrect target behaviors, missing or cropped pedestrians, and implausible scene compositions. After this filtering process, Attribute-Synthetic contains 1,333 high-quality synthetic samples. We provide further details on dataset construction and the prompts used for synthetic data generation in the supplementary material.

\noindent\textbf{(2) Surveillance Anomaly Datasets~\cite{AIHub-2,AIHub-multi,AIHub-senior}:}
To further enrich the visual coverage of rare safety-critical scenarios, we also curate real-world samples from surveillance anomaly datasets. The curated videos cover more than 60 indoor and outdoor environments and contain 1,431 video clips in total. We uniformly sample frames from each video at 1 FPS and apply {YOLO11x-pose}~\cite{yolo11_ultralytics} to extract pedestrian crops. We use a pose-based detector instead of a standard object detector because non-upright pedestrians, such as lying or fallen persons, are often missed by conventional detection models. This process initially yields 202,690 candidate pedestrian crops. We filter out erroneous detections, ambiguous crops, and crops containing multiple people. We further remove highly similar consecutive crops within each video to reduce near duplicates, resulting in 2,502 images.

\noindent\textbf{(3) Pedestrian Attribute Datasets~\cite{jin2024pedestrian, PETA, PA100k, RAP2}:} To cover common pedestrian appearances, we also integrate four pedestrian attribute datasets: MSP60K~\cite{jin2024pedestrian}, PETA~\cite{PETA}, PA100K~\cite{PA100k}, and RAP2~\cite{RAP2}. These datasets include both indoor and outdoor pedestrian images and provide diverse visual cues, including demographic attributes, clothing, accessories, carried objects, and viewpoints. In total, we collected 264,050 samples from these datasets.

EPAD contains 267,885 pedestrian images in total. To qualitatively examine the domain distribution, we visualize the image embedding distribution using 1,000 randomly sampled images from each dataset in~\cref{fig:dataset}(a). Rather than aiming to dominate the dataset in scale, Surveillance Anomaly and Attribute-Synthetic are introduced to expand the visual support for rare and safety-critical cases, such as non-upright poses, assistive-device usage, and threat-related object interactions, which are underrepresented in conventional pedestrian attribute datasets. Furthermore, as shown in Table~\ref{tab:dataset_statics}, EPAD provides a larger sample scale than UPAR~\cite{cormier2024upar}, a unified benchmark constructed from existing pedestrian attribute datasets. This enables evaluation under a more diverse set of pedestrian search queries.

\begin{table}[t]
\centering
\scriptsize
\setlength{\tabcolsep}{2pt}
\begin{tabularx}{\columnwidth}{
>{\raggedright\arraybackslash}p{0.18\columnwidth}
>{\centering\arraybackslash}X
>{\centering\arraybackslash}X
>{\centering\arraybackslash}X
}
\toprule
\multirow{2}{*}{\textbf{Category}} 
& \multirow{2}{*}{\textbf{Base} (26)} 
& \multicolumn{2}{c}{\textbf{Novel}} \\
\cmidrule(lr){3-4}
& & \textbf{In-Dist.} (23) & \textbf{Out-Dist.} (16) \\
\midrule

\textbf{Gender}
& male
& female
& -- \\
\midrule

\textbf{Age}
& young, adult
& senior
& -- \\
\midrule

\textbf{Hair}
& long hair, short hair, dark hair
& bald, light hair
& -- \\
\midrule

\textbf{Upper Clothing}
& short sleeve, jacket, coat, t-shirt, sweatshirt
& long sleeve, blouse, sweater, hoodie, blazer
& -- \\
\midrule

\textbf{Lower Clothing}
& long pants, short skirt, jeans
& shorts, long skirt, dress
& -- \\
\midrule

\textbf{Bag}
& briefcase, fanny pack
& backpack, shoulder bag, suitcase
& -- \\
\midrule

\textbf{Accessories}
& scarf, glasses, beanie, helmet, mask, headphones
& umbrella, muffler, sunglasses, hat, cap, earphones
& -- \\
\midrule

\textbf{Object Possession}
& using phone, holding child, box, \ggreen{gun}
& talking on phone, \ggreen{knife}
& -- \\
\midrule

\textbf{Mobility}
& --
& --
& motorcycle, bicycle, scooter, cart, \ggreen{stroller}, \ggreen{wheelchair}, \ggreen{cane}, \ggreen{crutches} \\
\midrule

\textbf{Action}
& --
& --
& smoking, eating, drinking, \ggreen{falling down}, \ggreen{lying down}, \ggreen{climbing fence}, \ggreen{drinking alcohol}, \ggreen{fighting} \\

\bottomrule
\end{tabularx}
\caption{Attribute split used for training and evaluation on EPAD. 
The 26 base attributes are used for training, whereas the 39 novel attributes are held out and used exclusively for evaluation. Novel attributes are considered in-distribution if their semantic category contains at least one base attribute, and out-of-distribution otherwise. As a result, 23 classes are categorized as in-distribution and 16 classes as out-of-distribution. \ggreen{Attributes highlighted in green} indicate identity-informative cues newly introduced in EPAD, serving as key evidence for target person retrieval.}
\label{tab:attribute_split}
\end{table}

\begin{table}[h] 
    \centering
    \resizebox{0.7\linewidth}{!}{%
    \begin{tabular}{rccc}
        \toprule
        Dataset & Attributes & Train & Test \\
        \toprule        
        MSP60K~\cite{jin2024pedestrian} & 57 &  36,300 & 23,822\\
        PA100K~\cite{PA100k} & 26 & 90,000 & 10,000 \\
        PETA~\cite{PETA} & 61 &  11,400 & 7,600 \\
        RAP2~\cite{RAP2} & 76 &  67,943 & 16,985 \\
        UPAR*~\cite{cormier2024upar} & 40 & 178,878 & 45,859 \\
        \midrule
        \textbf{EPAD (Ours)} & \textbf{65} & \textbf{187,519} & \textbf{80,366} \\
                \bottomrule
    \end{tabular}%
    }
    \caption{Comparison with existing pedestrian attribute datasets. EPAD provides a unified attribute vocabulary and a larger sample scale than prior unified benchmarks. * denotes a unified dataset constructed from existing pedestrian attribute datasets.}
    \vspace{-0.5em}
    \label{tab:dataset_statics}
    \vspace{-1em}
\end{table}

\subsection{Attribute Annotation}
EPAD defines a unified vocabulary of 65 pedestrian attributes, organized into 10 semantic categories: \textit{gender, age, hair, upper clothing, lower clothing, bag, accessories, object possession, mobility,} and \textit{action}. The complete attribute taxonomy is summarized in~\cref{tab:attribute_split}.
Specifically, we first inherit existing annotations from pedestrian attribute datasets whenever the target attributes are already available, including common demographic, clothing, and accessory. Beyond this label unification, EPAD expands the annotation space with newly introduced attributes, including \emph{gun}, \emph{knife}, \emph{stroller}, \emph{wheelchair}, \emph{cane}, \emph{crutches}, \emph{falling down}, \emph{lying down}, \emph{climbing fence}, \emph{drinking alcohol}, and \emph{fighting}. These attributes are particularly important for surveillance-oriented retrieval, where users may search for pedestrians based on safety-critical actions, assistive devices, or carried objects that are rarely annotated in conventional pedestrian attribute benchmarks. For attributes that are absent from the original datasets, we manually annotate the corresponding image-attribute pairs. In total, this process adds approximately 4M binary attribute annotations, enabling EPAD to evaluate pedestrian retrieval under a broader and more fine-grained attribute vocabulary.

\subsection{Sample and Attribute Splits}
We randomly shuffle all images and split them into 70\% training and 30\% testing sets, resulting in 187,519 training images and 80,366 test images as summarized in~\cref{tab:dataset_statics}.
For the attribute-level split, we partition the attribute classes into base and novel attributes, where only base attributes are used for training and novel attributes are reserved for evaluation, as shown in~\cref{tab:attribute_split}.
Novel attributes are further divided into in-distribution and out-of-distribution splits based on category-level exposure. 
Specifically, a novel attribute is in-distribution if its category includes at least one base attribute, and out-of-distribution otherwise. 
The in-distribution split measures generalization to unseen attributes within observed semantic categories, while the out-of-distribution split evaluates recognition of attributes from entirely unseen categories.
\section{Open-Attribute Person Retrieval}
\subsection{Task Definition}
We treat open-attribute person retrieval (OAPR) as a text-to-image search task. 
Regardless of whether an attribute is seen during training, the goal is to retrieve person images that match a given attribute query. 
Specifically, given a target attribute query $\pi=\{a_1,a_2,\cdots,a_r\}$, where $a_i\in\mathcal{A}$, the model retrieves the top-$K$ person images $\mathcal{X}=\{\mathcal{X}_1,\mathcal{X}_2,\cdots,\mathcal{X}_K\}$. The full attribute space $\mathcal{A}$ is divided into base and novel attributes:
\vspace{-0.5em}
\begin{align}
    \mathcal{A} = \mathcal{A}_{base} \cup \mathcal{A}_{novel}, 
    \quad
    \mathcal{A}_{novel} = \mathcal{A}_{in} \cup \mathcal{A}_{out},
\end{align}
\vspace{-0.5em}
where
\vspace{-0.5em}
\begin{align}
    \mathcal{A}_{base} \cap \mathcal{A}_{novel} = \varnothing,
    \quad
    \mathcal{A}_{in} \cap \mathcal{A}_{out} = \varnothing.
\end{align}
The model is trained only on the base attribute space $\mathcal{A}_{base}$ and evaluated on both base and novel attributes, i.e., $\mathcal{A}_{base}\cup\mathcal{A}_{novel}$. We further divide the novel attributes into in-distribution and out-of-distribution attributes.  The novel in-distribution attributes $\mathcal{A}_{in}$ belong to the same semantic attribute categories as $\mathcal{A}_{base}$, but consist of unseen attribute classes. 
In contrast, the novel out-of-distribution attributes $\mathcal{A}_{out}$ belong to entirely new attribute categories that do not overlap with those of $\mathcal{A}_{base}$ or $\mathcal{A}_{in}$. 
This split allows us to evaluate not only the model's ability to generalize to unseen classes within known categories, but also its transferability to attributes from previously unseen categories.

\subsection{Evaluation Metrics}
In practical open-attribute person retrieval, users usually inspect only a small set of top-ranked candidates rather than the entire ranked gallery. We therefore treat Rank@K and Precision@K as primary indicators of retrieval utility, as they directly measure whether relevant pedestrians appear early and how many of the inspected candidates satisfy the queried attributes. We additionally report mAP as a complementary full-ranking metric to evaluate the overall ordering quality across the complete retrieval list. 

\begin{figure*}[t]
  \centering
  \centering
  \begin{subfigure}{\textwidth}
    \includegraphics[width=\textwidth]{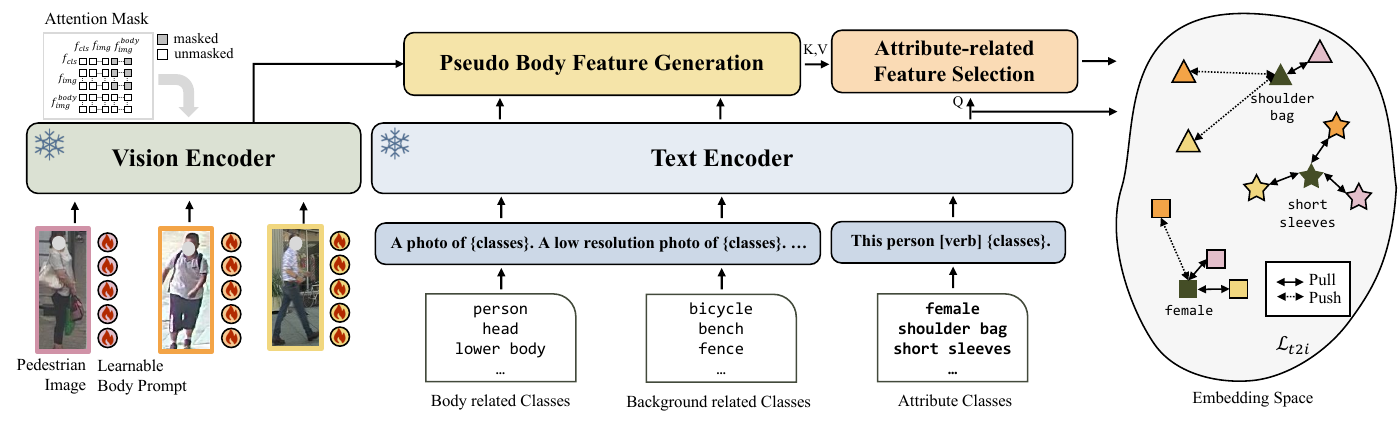}
    \caption{Overview of the training process of our proposed framework.}
    \label{fig:main-a}
  \end{subfigure}
  
  \begin{subfigure}{0.63\textwidth}
    \includegraphics[width=\linewidth]{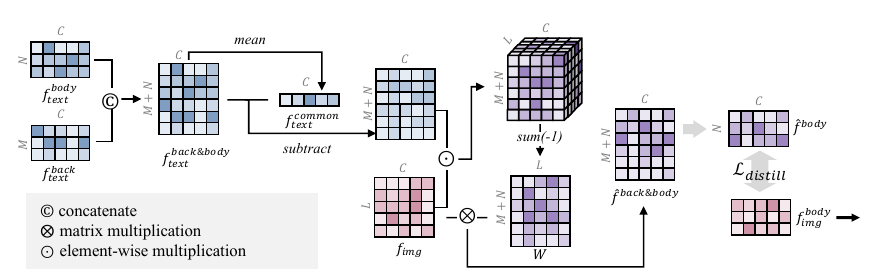}
    \caption{The process of the proposed pseudo body feature generation.}
    \label{fig:main-b}
  \end{subfigure}\hfill
  \begin{subfigure}{0.35\textwidth}
    \includegraphics[width=\linewidth]{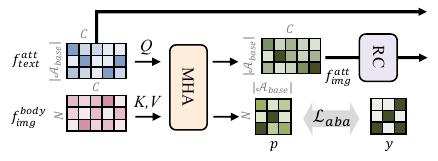}
    \caption{The process of the proposed attribute-related feature selection and gated residual correction.}
    \label{fig:main-c}
  \end{subfigure}
  
  \caption{Our framework consists of the vision and text encoders, a pseudo body feature generation module, and an attribute-related feature selection module. In the embedding space, the dark green triangle denotes the attribute text features $f_{text}^{att}$. Identical colors represent the same individual, and identical shapes denote the same attribute category.}
  \label{fig:main_combined}
  \vspace{-1.5em}
\end{figure*}

\subsection{GAP-CLIP}
\paragraph{Overview.} 

We observe that retrieval-critical attributes are often localized and body-part dependent (e.g., a gun, knife, cane, or crutches). 
Therefore, effective retrieval based on such cues requires the model to go beyond global image representation and localize the body region that provides visual evidence for the queried attribute.
Building on this observation, we propose GAP-CLIP, a CLIP-based framework that learns \textbf{G}ated \textbf{A}ttribute-aware body-\textbf{P}art representations for Open-Attribute Person Retrieval. 
GAP-CLIP leverages the open-vocabulary knowledge of CLIP while explicitly modeling body-part-aware visual representations. This design enables the model to focus on localized and discriminative visual cues that are critical for retrieval, while a gated residual connection preserves the original CLIP representation as a zero-shot anchor for unseen attributes.

As shown in~\cref{fig:main_combined}, our framework consists of a vision encoder, a text encoder, a pseudo-body feature generation module, and an attribute-related feature selection module. For the vision encoder, inspired by prior studies showing that diagonally prominent attention maps enhance local visual semantics by reducing patch-level noise~\cite{CLIPSurgery, AnomalyCLIP}, we replace the $V$ Q-K self-attention blocks with V-V self-attention in the vision encoder. Additionally, we apply a masking strategy to prevent the learnable prompts from interfering with the pre-trained representations. The learnable body prompt $Z=[z_1, z_2, ... ,z_N]$ and an input image $X$ are processed through the vision encoder to obtain the $f^{body} \in \mathbb{R}^{N\times C}$, and $f_{cls}\in \mathbb{R}^{C}$ and $f_{img}  \in \mathbb{R}^{L\times C}$ derived from $Z$ and $X$, respectively.

\begin{equation}
    [f_{cls},f_{img}, f_{img}^{body}] = \text{VisEnc}([X, Z]),
\end{equation}

For textual attribute embeddings, we define $N$ body-related classes (e.g., lower body, head) and $M$ background-related classes (e.g., bicycle, bench). We employ the prompt ensemble strategy~\cite{CLIP}, which combines various context templates (e.g., A photo of {class}, A low-resolution photo of {class}) to construct body and background prompts. By passing these prompts through the text encoder, we obtain the body and background text features $f_{text}^{body}\in \mathbb{R}^{N \times C}$ and $f_{text}^{back}\in \mathbb{R}^{M \times C}$. Subsequently, we construct a fixed natural-language description for each pedestrian attribute. Specifically, each attribute is represented in the form of ``This person [verb] [attribute class]'', such as ``This person is wearing a hoodie'', ``This person is holding a gun'', or ``This person is lying down''. These attribute sentences are encoded by the CLIP text encoder to obtain the attribute text features
$f_{text}^{att}=[f_{text}^{att(1)},...,f_{text}^{att(|\mathcal{A}_{base}|)}]\in \mathbb{R}^{|\mathcal{A}_{base}| \times C}$. Then $f_{img}, f_{text}^{body}$ and $f_{text}^{back}$ pass through the pseudo body feature generation module to produce the pseudo body feature, which is distilled into $f_{img}^{body}$. 

The proposed attribute-related feature selection module takes $f_{text}^{att}$ as the query and $f_{img}^{body}$ as the key and value, deriving $f^{att} \in \mathbb{R}^{|\mathcal{A}_{base}|\times C}$ with enhanced attribute-specific visual representations. For effective training, we introduce an attribute part association loss to enhance the model’s ability to learn the body parts associated with each attribute. Finally, as shown in~\cref{fig:main-a}, we compute the similarity between the batch of attribute-conditioned visual features and the text features of each attribute, pulling positive pairs closer and pushing negative pairs apart via a text-to-image contrastive loss. 

\vspace{-1.5em}
\paragraph{Pseudo Body Feature Generation.}
\label{sec:method-pseudo}
Although CLIP provides strong open-vocabulary image-text representations, its global image feature is not explicitly aware of pedestrian body structure. Since regional body information is crucial for person retrieval and attribute recognition~\cite{SOLIDER,ddddpoar,PromptPAR,park2025vita}, we aim to inject body-part awareness into CLIP representations while preserving their transferable semantic knowledge.
However, explicit body-part annotations are often unavailable in pedestrian attribute datasets, and surveillance images frequently contain background clutter that can interfere with local attribute modeling. To obtain body-aware supervision without additional part labels, we generate pseudo body features from the CLIP image-text embedding space, as shown in~\cref{fig:main-b}. Inspired by~\cite{CLIPSurgery}, we construct text features for both body-related classes and background-related classes, and use them to estimate which image patches are associated with body regions. Specifically, we concatenate $f_{text}^{back}$ and $f_{text}^{body}$ to obtain $f_{text}^{back\&body}$, and compute their mean feature vector $f^{common}_{text}$. This common feature represents the class-irrelevant component shared by both body and background prototypes. By subtracting it from each class prototype, the remaining textual directions emphasize class-specific cues and make the resulting patch responses more separable between body and background regions. We then compute the patch-level response map $W$ by measuring the similarity between the image patch features $f_{img}$ and the debiased textual prototypes. The response map highlights image patches that are relevant to each body or background class. Finally, we aggregate the patch features according to $W$ to obtain the pseudo body/background features $\hat{f}^{back\&body}$ as follows:


\vspace{-1.3em}
\begin{align}
    W = {f_{img}}^T(f^{back\&body}_{text} - f^{common}_{text})
\end{align}
\begin{align}
    {\hat{f}}^{back\& body}=Wf_{img} 
\end{align}


We select only  body part proportion $\hat{f}^{body}$ from ${\hat{f}}^{back\& body}$, which serves as the pseudo body feature. The pseudo feature supervises $f_{img}^{body}$ via a cosine distance loss:
\begin{align}
\mathcal{L}_{distill} = 1 - \mathrm{sim}(f^{body}_{img}, \hat{f}^{body}),
\end{align}
where $\mathrm{sim}(\cdot,\cdot)$ denotes the cosine similarity.

\begin{table*}[t]
\centering
\resizebox{\linewidth}{!}{
\begin{tabular}{llcccccccccccccccccccc}
\toprule
\multirow{3}{*}{Category}
& \multirow{3}{*}{Method}
& \multicolumn{5}{c}{\multirow{2}{*}{Base}}
& \multicolumn{10}{c}{Novel}
& \multicolumn{5}{c}{\multirow{2}{*}{All}} \\
\cmidrule(lr){8-17}
& & & & &  & &\multicolumn{5}{c}{In-Dist}
& \multicolumn{5}{c}{Out-Dist}
& \\
\cmidrule(lr){3-7}
\cmidrule(lr){8-12}
\cmidrule(lr){13-17}
\cmidrule(lr){18-22}
&
&
R@1 & R@5 & P@1 & P@5 & mAP
&
R@1 & R@5 & P@1 & P@5 & mAP
&
R@1 & R@5 & P@1 & P@5 & mAP
&
R@1 & R@5 & P@1 & P@5 & mAP \\
\midrule
\multirow{4}{*}{VLM}
& CLIP~\cite{CLIP}          &  38.46 &  69.23 &  38.46 &  45.38 &  25.10 & 
 39.13 &  \best{78.26} &  39.13 &  33.91 &  13.11 & 
 43.75 &  75.00 &  43.75 &  45.00 &  26.68 &
 40.00 &  \second{73.85} &  40.00 &  41.23 &  21.25 \\
& OpenCLIP~\cite{OPENCLIP}      &  50.00 &  \second{73.08} &  50.00 &  47.69 &  24.08 &  \best{43.48} &  \best{78.26} &  \best{43.48} &  39.13 &  14.27 & 
 \second{56.25} &  62.50 &  \second{56.25} &  48.75 &  27.13 &  \second{49.23} &  72.31 &  \second{49.23} &  44.92 &  21.36 \\
& SigLIP~\cite{SigLIP}        &  11.54 &  26.92 &  11.54 &  16.15 &  19.20 &
 4.35 &  26.09 &  4.35 &  6.09 &  8.17 & 
 0.00 &  0.00 &  0.00 &  0.00 &  0.17 & 
 6.15 &  20.00 &  6.15 &  8.62 &  10.61 \\
& BLIP~\cite{BLIP}         &
 46.15 &  69.23 &  46.15 &  50.00 &  30.87 
&  43.48 &  73.91 &  \best{43.48} &  \best{42.61} &  21.96 & 
 50.00 &  \second{81.25} &  50.00 &  \second{55.00} &  \best{34.35} & 
 46.15 &  \second{73.85} &  46.15 &  48.62 &  \best{28.57}  \\
\midrule
\multirow{2}{*}{OD}
& GroundingDINO~\cite{groundingdino} & 26.92 & 30.77 & 26.92 & 21.54 & 19.79 & 
4.35 & 26.09 & 4.35 & 6.96 & 8.51 & 
0.00 & 0.00 & 0.00 & 0.00 & 0.20 & 
12.31 & 21.54 & 12.31 & 11.08 & 10.98 \\
& YOLO-World~\cite{yoloworld}    &   34.62 &  65.38 &  34.62 &  33.08 &  21.94 &  30.43 &  34.78 &  30.43 &  22.61 &  10.28 &  31.25 &  37.50 &  31.25 &  20.00 &  3.48 &  32.31 &  47.69 &  32.31 &  26.15 &  13.27 \\
\midrule
\multirow{1}{*}{PAR}
& POAR~\cite{ddddpoar}    & 26.92 & 57.69& 26.92 & 30.77 & 20.34
& 21.74 & 60.87 & 21.74 & 26.09 & 11.87
& 12.50 & 12.50 & 12.50 & 8.75 & 5.86
& 21.54 & 33.85 & 21.54 & 23.08 & 13.77 \\
\midrule
\multirow{2}{*}{TBPR}
& IRRA~\cite{irra}          &  73.08 &  \best{84.62} &  \second{73.08} &  \best{73.08} &  \best{43.39} &
 26.09 &  65.22 &  26.09 &  37.39 &  \best{22.55} &
 43.75 & 56.25 &  43.75 &  31.25 &  10.73 &
 \second{49.23} &  70.77 &  \second{49.23} &  \second{50.15} &  \second{27.98} \\
& PLOT~\cite{park2024plot}          
&  \best{76.92} &  \best{84.62} &  \best{76.92} &  \second{66.92} &  \second{38.86} &
 34.78 &  \best{78.26} &  34.78 &  \second{40.87} &  \second{17.91} & 
 18.75 &  37.50 &  18.75 & 20.00 & 4.08  &
 47.69 &  70.77 &  47.69 &  46.15 &  22.89 \\
\midrule
\rowcolor{blue!10}\textbf{OAPR} & \textbf{GAP-CLIP (Ours)}           &  \textbf{\second{57.69}} &  \textbf{\best{84.62}} &\textbf{\second{57.69}} &  \textbf{62.31} &  \textbf{31.96 }
&  \textbf{\best{43.48}} &  \textbf{\second{73.91}} &  \textbf{\best{43.48}} &  \textbf{\second{40.87}} &  \textbf{13.16 }
& \textbf{ \best{62.50}} &  \textbf{\best{87.50} }&  \textbf{\best{62.50}} &  \textbf{\best{66.25}} &  \textbf{\second{28.31} }
&  \textbf{\best{53.85}} &  \textbf{\best{81.54}} &  \textbf{\best{53.85}} &  \textbf{\best{55.38}} &  \textbf{\best{28.57}} \\
\bottomrule
\end{tabular}
}
\caption{Comparison with existing SOTA models. For each split, we report R@K and P@K, where R and P denote Rank and Precision, respectively, along with mAP. Best and second-best results are marked in \best{blue} and \second{orange}, respectively.
}
\label{tab:comparison}
\end{table*}

\vspace{-1em}
\paragraph{Attribute-Related Feature Selection.}
\label{sec:method-select}
Given that certain attributes exhibit spatial dependencies across multiple body parts (e.g., long hair extends across the head and upper body, and a dress spans both upper and lower body), we introduce an attribute-related feature selection module to adaptively focus on the most relevant regions for each attribute through an attention mechanism, as shown in~\cref{fig:main-c}. Given that $f_{text}^{att}$ and $f_{img}^{body}$, attention weight $p\in \mathbb{R}^{N \times |\mathcal{A}_{base}|}$ and the attribute-conditioned visual features $f^{att}_{img} = [f^{att(1)}_{img}, ..., f^{att(|\mathcal{A}_{base}|)}_{img}]\in \mathbb{R}^{|\mathcal{A}_{base}| \times C}$ computed as:
\begin{align}
p = \text{Softmax}(\frac{f_{text}^{att}W^Q (f_{img}^{body}W^K)^\top}{\sqrt{d_k}} )
\end{align}
\begin{align}
\tilde{f}^{att}_{img} = p(f_{img}^{body}W^V)
\end{align}
where $W^Q,W^K,W^V$ represent the query, key, and value projection matrices, respectively, incorporating the multi-head mechanism.

Since the body tokens are learned without explicit part annotations, the attention weights may not be well aligned with semantically relevant body regions during training. To guide the attribute-related feature selection module, we introduce an attribute-body association (ABA) loss that provides weak supervision on the expected body region for each attribute. Specifically, each attribute is assigned to one or more relevant body parts; for example, ``short-sleeved t-shirt'' is associated with the upper-body region, while ``long hair'' is associated with the head region. The ABA loss encourages the attention distribution of each attribute to focus on its corresponding body regions:
\begin{align}
    \mathcal{L}_{aba}
    =
    {1 \over |\mathcal{A}_{base}|}
    \sum_{i=1}^{|\mathcal{A}_{base}|}
    \sum_{j=1}^{N}
    -y^{(i)}_j \log p^{(i)}_j,
\end{align}
where $y^{(i)}_j$ indicates whether the $i$-th attribute is associated with the $j$-th body part.

\vspace{-1em}
\paragraph{Gated Residual Correction.}
Although the attribute-related feature selection module provides attribute-specific visual representations, directly replacing the final image representation with the cross-attention features can distort the pre-trained vision-language embedding space. This is particularly undesirable in the open-attribute setting, where the model should preserve the zero-shot knowledge of CLIP for attributes that are not observed during training. To mitigate this issue, we introduce a gated residual connection. Let $\tilde{f}^{att}_{img}\in\mathbb{R}^{|\mathcal{A}_{base}|\times C}$ denote the normalized output of the attribute-related cross-attention module, and let $f_{cls}\in\mathbb{R}^{C}$ be the normalized CLIP class-token feature. We use $f_{cls}$ as a global image-level anchor that retains the original CLIP representation. For each attribute, the final attribute-conditioned visual feature is computed as
\begin{align}
    \alpha = \sigma(\gamma),
\end{align}
\begin{align}
    f^{att}_{img}
    =
    f_{cls} + \alpha \tilde{f}^{att}_{img}
    ,
\end{align}
where $\gamma$ is a learnable scalar initialized by a hyperparameter and $\sigma(\cdot)$ denotes the sigmoid function.
This formulation treats the CLIP class token as a frozen zero-shot anchor and allows the cross-attention feature to act as an attribute-specific residual correction. The gate $\alpha$ controls the strength of this correction, and the learnable gate controls the overall magnitude of the attribute-specific correction. We initialize $\gamma=-3.0$, yielding $\alpha=\sigma(\gamma)\approx0.047$. Therefore, this process stabilizes training and reduces overfitting to base attributes, while maintaining the transferability of CLIP representations to novel attributes.



Finally, we calculate the similarity between the batch of the attribute-conditioned visual features and the text attribute feature. 
To achieve this, we propose a text-to-image contrastive loss to effectively align image features with text features, ensuring a more precise and semantically meaningful feature correspondence.
\begin{align}
    \mathcal{L}_{t2i} = \sum_{i=1}^{|\mathcal{A}_{base}|} -\log \biggl( \frac{S_B^{i+}}{S_B^{i+} + w_{neg} S_B^{i-}} \biggr)
\end{align}
\vspace{-1em}

\begin{align}
    S_B^{i+} = \exp({I_{B}^{i+} f_{text}^{att(i)}}/\tau),\;S_B^{i-} = \exp({I_{B}^{i-} f_{text}^{att(i)}}/\tau)
\end{align}
where $f_{text}^{att(i)}$ refers to the $i$-th text attribute feature. $I_{B}^{i+}$ denotes the set of positive pairs between the batch of $i$-th attribute-conditioned visual features $f^{att(i)}_{img}$ and the corresponding $f_{text}^{att(i)}$, while $I_{B}^{i-}$ denotes the set of negative pairs.


Overall, during the training phase, we employ three loss functions for the total loss function given:
\begin{align}
    \mathcal{L}_{train}=\mathcal{L}_{t2i} + \lambda_{distill}\mathcal{L}_{distill} + \lambda_{aba}\mathcal{L}_{aba}
\end{align}
where $\lambda_{distill}, \lambda_{aba}$ are the hyperparameters.

\section{Experiment}
\begin{figure*}[t]
    \centering
    \includegraphics[width=\textwidth]{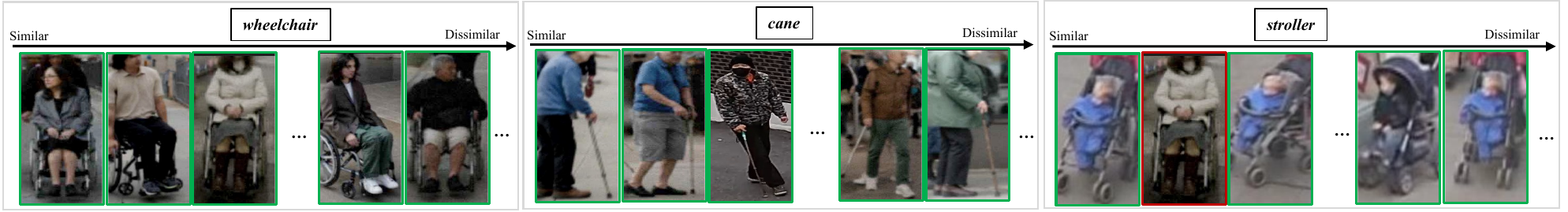}
    \caption{Qualitative retrieval results on out-of-distribution attributes. Retrieved images are arranged from more similar to less similar according to the predicted image-text similarity. {Green boxes} indicate {correctly retrieved images} that match the queried attribute, while {red boxes} indicate {incorrect retrievals}. The results show that our model can successfully retrieve pedestrians with unseen attributes, while also revealing challenging cases caused by visually similar but semantically different cues.}
    \label{fig:quan}
\end{figure*}
\begin{figure}[t] 
    \centering
    \includegraphics[width=\linewidth]{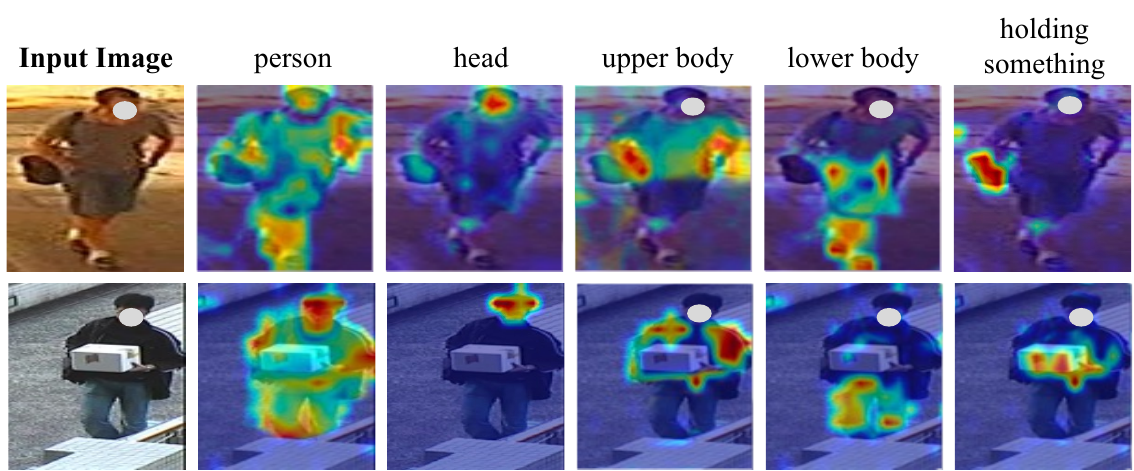} 
    \caption{Visualization demonstrating the effectiveness of the pseudo-body generation module. }
    \label{fig:attentionmap}
    
\end{figure}

\paragraph{Implementation Details.}
We adopt pre-trained CLIP (ViT-B/16)~\cite{CLIP} and freeze both image and text encoders. All training and evaluation experiments are conducted on four NVIDIA A6000 GPUs. The number of distinct body features ($N$) is set to $5$, and we use the following body part classes: \textit{``person"}, \textit{``head"}, \textit{``upper body"}, \textit{``lower body"}, \textit{``holding something"}. $V$ is set to 6, and the temperature parameter $\tau$ is set to $0.1$, and the weighting factor for negative samples, $w_{\text{neg}}$, is set to $1$. The hyperparameters $\lambda_{distill}$ and $\lambda_{aba}$ are set to $0.1$ and $0.02$, respectively. Further configuration details are provided in the supplementary material. 

\paragraph{Quantitative Results.}
Table~\ref{tab:comparison} presents the comparison with VLMs (CLIP~\cite{CLIP}, OpenCLIP~\cite{OPENCLIP}, SigLIP~\cite{SigLIP}, BLIP~\cite{BLIP}), open-vocabulary object detection models (GroundingDINO~\cite{groundingdino}, YOLO-World~\cite{yoloworld}), an open-vocabulary PAR model (POAR~\cite{POAR}), and text-based person retrieval models (IRRA~\cite{irra}, PLOT~\cite{park2024plot}). 
Table 3 reveals a clear distinction between closed/base-attribute retrieval and open-attribute retrieval. On base attributes, while existing TBPR models achieve strong performance on base attributes, their retrieval accuracy decreases substantially when the query attributes belong to unseen semantic categories. This indicates that models trained for identity-centric text-based person retrieval can effectively exploit familiar textual descriptions, but are less robust to category-level open attributes that are critical in practical surveillance search. GAP-CLIP shows clear advantages in this open-attribute setting. On `Out-Dist' attributes, where the queried attributes come from entirely unseen attribute categories, GAP-CLIP achieves superior performance. This demonstrates that the proposed framework is effective at retrieving pedestrians based on retrieval-critical cues that are not covered by the base attribute space. Notably, GAP-CLIP improves the retrieval quality, which is particularly important for real-world search scenarios where operators typically inspect only a limited number of top results. The advantage of GAP-CLIP is also evident in the `All' setting, which evaluates retrieval over the full attribute space. These results suggest that explicitly modeling attribute-relevant body regions enables the model to capture localized and discriminative cues more effectively than global VLM matching, open-vocabulary detection, or identity-centric person retrieval baselines.

\vspace{-1em}
\paragraph{Qualitative Results.}
Fig.~\ref{fig:quan} shows qualitative retrieval results for out-of-distribution attributes, including \textit{wheelchair}, \textit{cane}, and \textit{stroller}. Overall, GAP-CLIP retrieves visually consistent results that match the queried attributes, even though these attributes belong to unseen semantic categories during training. 
In particular, the model successfully captures localized retrieval-critical cues, such as assistive devices and wheeled objects, which often occupy only a small region of the pedestrian image. 
These examples further support the effectiveness of our body-part-aware attribute modeling for open-attribute person retrieval.

In addition, as shown in~\cref{fig:attentionmap}, we visualize the activation map for the ``person", ``head", ``upper body", ``lower body" and ``holding something" of $W$ in the pseudo body feature generation module to demonstrate its effectiveness. The maps are color-coded, where red indicates a high response. Our proposed module effectively highlights the regions of interest for each body part without training cost. Specifically, the response for ``person'' broadly covers the entire pedestrian region while suppressing much of the background, indicating that the generated response is person-centric. In contrast, the responses for ``head'', ``upper body'', and ``lower body'' are concentrated on their corresponding body regions, showing that the textual prototypes can induce semantically localized patch responses. Moreover, the activation for ``holding something'' highlights regions around the hands and carried objects, which are often small but important cues for attribute retrieval. These results suggest that the proposed pseudo-body generation module can provide meaningful body-aware supervision without requiring explicit body-part annotations.
\vspace{-0.5em}
\section{Conclusion}
\label{sec:conclusion}
\vspace{-0.5em}
In this paper, we introduced Open-Attribute Person Retrieval (OAPR), a practical retrieval setting that aims to find pedestrians matching attribute queries, including rare and previously unseen visual concepts. To support this task, we constructed EPAD, a large-scale benchmark that expands existing pedestrian attribute annotations with retrieval-critical attributes. We further proposed GAP-CLIP, a lightweight CLIP-based framework that learns gated attribute-aware body-part representations for open-attribute retrieval. Extensive experiments on EPAD show that GAP-CLIP improves retrieval performance over vision-language, open-vocabulary detection, open-vocabulary PAR, and text-based person retrieval baselines, particularly on novel and out-of-distribution attributes. We hope that OAPR and EPAD will facilitate future research on attribute-centric person search in realistic surveillance scenarios.

{
    \small
    \bibliographystyle{ieeenat_fullname}
    \bibliography{main}

@String(CVPR= {IEEE Conf. Comput. Vis. Pattern Recog.})

@String(ICIP = {IEEE Int. Conf. Image Process.})

@String(CVPR  = {CVPR})

@String(ICIP  = {ICIP})

@misc{ideogram-4-2026,
    author={Ideogram AI},
    title={{Ideogram 4}},
    year={2026},
    howpublished={\url{https://ideogram.ai/blog/ideogram-4.0/}},
}

@misc{AIHub-senior,
    author={{The Open AI Dataset Project (AI-Hub, S. Korea)}},
    title={{Anomalous Behaviors in Seniors}},
    year={2020},
    howpublished={\url{www.aihub.or.kr}},
}

@misc{AIHub-multi,
    author={{The Open AI Dataset Project (AI-Hub, S. Korea)}},
    title={{Multi-Angle CCTV Data for Public Safety}},
    year={2025},
    howpublished={\url{www.aihub.or.kr}},
}

@misc{AIHub-2,
    author={{The Open AI Dataset Project (AI-Hub, S. Korea)}},
    title={{CCTV Video Dataset for Monitoring Key Facilities and Illegal Activities in Parks}},
    year={2021},
    howpublished={\url{www.aihub.or.kr}},
}

@inproceedings{CLIP,
  title={Learning transferable visual models from natural language supervision},
  author={Radford, Alec and Kim, Jong Wook and Hallacy, Chris and Ramesh, Aditya and Goh, Gabriel and Agarwal, Sandhini and Sastry, Girish and Askell, Amanda and Mishkin, Pamela and Clark, Jack and others},
  booktitle={International conference on machine learning},
  pages={8748--8763},
  year={2021},
  organization={PMLR}
}

@inproceedings{BLIP,
  title={Blip: Bootstrapping language-image pre-training for unified vision-language understanding and generation},
  author={Li, Junnan and Li, Dongxu and Xiong, Caiming and Hoi, Steven},
  booktitle={International conference on machine learning},
  pages={12888--12900},
  year={2022},
  organization={PMLR}
}

@inproceedings{ALIGN,
  title={Scaling up visual and vision-language representation learning with noisy text supervision},
  author={Jia, Chao and Yang, Yinfei and Xia, Ye and Chen, Yi-Ting and Parekh, Zarana and Pham, Hieu and Le, Quoc and Sung, Yun-Hsuan and Li, Zhen and Duerig, Tom},
  booktitle={International conference on machine learning},
  pages={4904--4916},
  year={2021},
  organization={PMLR}
}

@inproceedings{OPENCLIP,
  title={Reproducible scaling laws for contrastive language-image learning},
  author={Cherti, Mehdi and Beaumont, Romain and Wightman, Ross and Wortsman, Mitchell and Ilharco, Gabriel and Gordon, Cade and Schuhmann, Christoph and Schmidt, Ludwig and Jitsev, Jenia},
  booktitle={Proceedings of the IEEE/CVF Conference on Computer Vision and Pattern Recognition},
  pages={2818--2829},
  year={2023}
}

@inproceedings{SOLIDER,
  title={Beyond Appearance: a Semantic Controllable Self-Supervised Learning Framework for Human-Centric Visual Tasks},
  author={Chen, Weihua and Xu, Xianzhe and Jia, Jian and Luo, Hao and Wang, Yaohua and Wang, Fan and Jin, Rong and Sun, Xiuyu},
  booktitle={Proceedings of the IEEE/CVF Conference on Computer Vision and Pattern Recognition},
  pages={15050--15061},
  year={2023}
}

@misc{CLIPSurgery,
      title={CLIP Surgery for Better Explainability with Enhancement in Open-Vocabulary Tasks}, 
      author={Yi Li and Hualiang Wang and Yiqun Duan and Xiaomeng Li},
      year={2023},
      eprint={2304.05653},
      archivePrefix={arXiv},
      primaryClass={cs.CV},
      url={https://arxiv.org/abs/2304.05653}, 
}

@inproceedings{AnomalyCLIP,
  title={AnomalyCLIP: Object-agnostic Prompt Learning for Zero-shot Anomaly Detection},
  author={Zhou, Qihang and Pang, Guansong and Tian, Yu and He, Shibo and Chen, Jiming},
  booktitle={The Twelfth International Conference on Learning Representations},
  year={2023}
}

@inproceedings{cormier2024upar,
  title={Upar challenge 2024: Pedestrian attribute recognition and attribute-based person retrieval-dataset, design, and results},
  author={Cormier, Mickael and Specker, Andreas and Junior, Julio and Jacques, CS and Moritz, Lennart and Metzler, J{\"u}rgen and Moeslund, Thomas B and Nasrollahi, Kamal and Escalera, Sergio and Beyerer, J{\"u}rgen},
  booktitle={Proceedings of the IEEE/CVF Winter Conference on Applications of Computer Vision},
  pages={359--367},
  year={2024}
}

@article{RAP2,
title={A richly annotated pedestrian dataset for person retrieval in real surveillance scenarios},
author={Li, Dangwei and Zhang, Zhang and Chen, Xiaotang and Huang, Kaiqi},
journal={IEEE transactions on image processing},
volume={28},
number={4},
pages={1575--1590},
year={2019},
publisher={IEEE}
}

@inproceedings{li2017person,
  title={Person search with natural language description},
  author={Li, Shuang and Xiao, Tong and Li, Hongsheng and Zhou, Bolei and Yue, Dayu and Wang, Xiaogang},
  booktitle={Proceedings of the IEEE conference on computer vision and pattern recognition},
  pages={1970--1979},
  year={2017}
}

@inproceedings{zhu2021dssl,
  title={Dssl: Deep surroundings-person separation learning for text-based person retrieval},
  author={Zhu, Aichun and Wang, Zijie and Li, Yifeng and Wan, Xili and Jin, Jing and Wang, Tian and Hu, Fangqiang and Hua, Gang},
  booktitle={Proceedings of the 29th ACM international conference on multimedia},
  pages={209--217},
  year={2021}
}

@inproceedings{PA100k,
    Title = {HydraPlus-Net: Attentive Deep Features for Pedestrian Analysis},
    author = {Liu, Xihui and Zhao, Haiyu and Tian, Maoqing and Sheng, Lu and Shao, Jing and Yan, Junjie and Wang, Xiaogang},
    Booktitle = {Proceedings of the IEEE international conference on computer vision},
    pages={1--9},
    year={2017}
}

@inproceedings{PETA,
  title={Pedestrian attribute recognition at far distance},
  author={Deng, Yubin and Luo, Ping and Loy, Chen Change and Tang, Xiaoou},
  booktitle={Proceedings of the 22nd ACM international conference on Multimedia},
  pages={789--792},
  year={2014}
}

@article{PromptPAR,
  author={Wang, Xiao and Jin, Jiandong and Li, Chenglong and Tang, Jin and Zhang, Cheng and Wang, Wei},
  title={Pedestrian Attribute Recognition via CLIP-Based Prompt Vision-Language Fusion}, 
  journal={IEEE Transactions on Circuits and Systems for Video Technology}, 
  year={2025}}

@inproceedings{ddddpoar,
  title     = {POAR: Towards Open Vocabulary Pedestrian Attribute Recognition},
  author    = {Yue Zhang and Suchen Wang and Shichao Kan and Zhenyu Weng and Yigang Cen and Yappeng Tan},
  booktitle = {Proceedings of the 31st ACM International Conference on Multimedia (MM '23)},
  year      = {2023},
  doi       = {10.1145/3581783.3611719}
}

@inproceedings{park2025vita,
  title={ViTA-PAR: visual and textual attribute alignment with attribute prompting for pedestrian attribute recognition},
  author={Park, Minjeong and Park, Hongbeen and Kim, Jinkyu},
  booktitle={2025 IEEE International Conference on Image Processing (ICIP)},
  pages={31--36},
  year={2025},
  organization={IEEE}
}

@article{jin2024pedestrian,
  title={Pedestrian Attribute Recognition: A New Benchmark Dataset and A Large Language Model Augmented Framework},
  author={Jin, Jiandong and Wang, Xiao and Zhu, Qian and Wang, Haiyang and Li, Chenglong},
  journal={arXiv preprint arXiv:2408.09720},
  year={2024}
}

@software{yolo11_ultralytics,
  author = {Glenn Jocher and Jing Qiu},
  title = {Ultralytics YOLO11},
  version = {11.0.0},
  year = {2024},
  url = {https://github.com/ultralytics/ultralytics},
  orcid = {0000-0001-5950-6979, 0000-0003-3783-7069},
  license = {AGPL-3.0}
}

@inproceedings{park2024plot,
  title={Plot: Text-based person search with part slot attention for corresponding part discovery},
  author={Park, Jicheol and Kim, Dongwon and Jeong, Boseung and Kwak, Suha},
  booktitle={European Conference on Computer Vision},
  pages={474--490},
  year={2024},
  organization={Springer}
}

@inproceedings{SigLIP,
  title={Sigmoid loss for language image pre-training},
  author={Zhai, Xiaohua and Mustafa, Basil and Kolesnikov, Alexander and Beyer, Lucas},
  booktitle={Proceedings of the IEEE/CVF international conference on computer vision},
  pages={11975--11986},
  year={2023}
}

@article{groundingdino,
  title={Grounding dino: Marrying dino with grounded pre-training for open-set object detection},
  author={Liu, Shilong and Zeng, Zhaoyang and Ren, Tianhe and Li, Feng and Zhang, Hao and Yang, Jie and Li, Chunyuan and Yang, Jianwei and Su, Hang and Zhu, Jun and others},
  journal={arXiv preprint arXiv:2303.05499},
  year={2023}
}

@inproceedings{yoloworld,
  title={YOLO-World: Real-Time Open-Vocabulary Object Detection},
  author={Cheng, Tianheng and Song, Lin and Ge, Yixiao and Liu, Wenyu and Wang, Xinggang and Shan, Ying},
  booktitle={Proc. IEEE Conf. Computer Vision and Pattern Recognition (CVPR)},
  year={2024}
}

@inproceedings{irra,
  title={Cross-Modal Implicit Relation Reasoning and Aligning for Text-to-Image Person Retrieval},
  author={Jiang, Ding and Ye, Mang},
  booktitle={IEEE International Conference on Computer Vision and Pattern Recognition (CVPR)},
  year={2023},
}

@article{ding2021semantically,
  title={Semantically self-aligned network for text-to-image part-aware person re-identification},
  author={Ding, Zefeng and Ding, Changxing and Shao, Zhiyin and Tao, Dacheng},
  journal={arXiv preprint arXiv:2107.12666},
  year={2021}
}
}

\end{document}